\documentclass[manuscript,screen]{acmart}

\AtBeginDocument{%
  \providecommand\BibTeX{{%
    \normalfont B\kern-0.5em{\scshape i\kern-0.25em b}\kern-0.8em\TeX}}}

\setcopyright{rightsretained}

\begin{document}
\copyrightyear{2021}
\acmYear{2021}
\acmConference[COMPASS '21]{ACM SIGCAS Conference on Computing and Sustainable Societies (COMPASS)}{June 28-July 2, 2021}{Virtual Event, Australia}
\acmBooktitle{ACM SIGCAS Conference on Computing and Sustainable Societies (COMPASS) (COMPASS '21), June 28-July 2, 2021, Virtual Event, Australia}\acmDOI{10.1145/3460112.3471966}
\acmISBN{978-1-4503-8453-7/21/06}

\title{Species Distribution Modeling for Machine Learning Practitioners: A Review}

\author{Sara Beery}
\authornote{Equal contribution.}
\email{sbeery@caltech.edu}
\author{Elijah Cole}
\authornotemark[1]
\email{ecole@caltech.edu}
\affiliation{%
  \institution{California Institute of Technology}
  \country{USA}
}

\author{Joseph Parker}
\affiliation{%
  \institution{California Institute of Technology}
  \country{USA}
}

\author{Pietro Perona}
\affiliation{%
 \institution{California Institute of Technology}
 \country{USA}
 }
 
 \author{Kevin Winner}
\affiliation{%
  \institution{Yale University}
  \country{USA}
  }

\begin{abstract}

Conservation science depends on an accurate understanding of what's happening in a given ecosystem.
How many species live there?
What is the makeup of the population?
How is that changing over time?
Species Distribution Modeling (SDM) seeks to predict the spatial (and sometimes temporal) patterns of \textit{species occurrence}, i.e. where a species is likely to be found. 
The last few years have seen a surge of interest in applying powerful machine learning tools to challenging problems in ecology \cite{AI_Animal_ReID, AI_CVWC_ICCV, FGVC}. 
Despite its considerable importance, SDM has received relatively little attention from the computer science community. 
Our goal in this work is to provide computer scientists with the necessary background to read the SDM literature and develop ecologically useful ML-based SDM algorithms. 
In particular, we introduce key SDM concepts and terminology, review standard models, discuss data availability, and highlight technical challenges and pitfalls.
\end{abstract}



\begin{teaserfigure}
      \centering
    \includegraphics[width=0.97\textwidth]{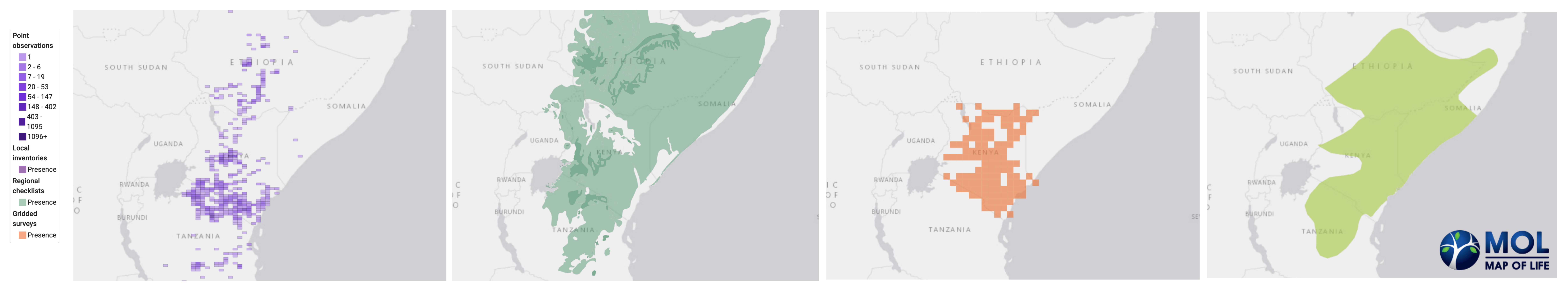}
    \caption{
    Species distribution models describe the relationship between environmental conditions and (actual or potential) species presence. However, the link between the environment and species distribution data can be complex, particularly since distributional data comes in many different forms. Above are four different sources of distribution data for the \textit{Von Der Decken's Hornbill} \cite{hornbill}: (from left to right) raw point observations, regional checklists, gridded ecological surveys, and data-driven expert range maps. All images are from Map of Life \cite{jetz2012integrating}.}
    \label{fig:distributions_comparison}
\end{teaserfigure}

\maketitle

\section{Introduction}

\begin{table*}[]
    \centering
    \begin{tabular}{r|c|l}
        \textbf{Data collection method} & \textbf{Example} & \textbf{Observation type}\\
        \hline 
        Community science observations & iNaturalist & Presence-only \\
        Community science checklists & eBird & Presence-absence \\
        Static sensors & Camera traps & Presence-absence \\
        Sample collection & Insect trapping & Presence-absence \\
        Expert field surveys & Line transects & Presence-absence \\
        Historic records, natural history collections & Herbarium sheets & Presence-only
    \end{tabular}
    \caption{Sources of species observation data. Each of these examples represents a method of collecting or accessing observations of different species. 
    One important distinction is whether the observations are \textit{presence-only} or \textit{presence-absence}.
    Presence-only data consists of locations where a species has been sighted. 
    Presence-absence data also includes locations where a species was checked for but not observed.}
    \label{tab:data_collection_methods}
\end{table*}

Ecological research helps us to understand ecosystems and how they respond to climate change, human activity, and conservation policies.
Much of this work starts by deploying networks of sensors (often cameras or microphones) to monitor the organisms living in a fixed study area.
Ecologists must then invest significant effort to filter, label, and analyze this data. 
This step is often a bottleneck for ecological research. 
For example, it can take years for scientists to process and interpret a single season of data from a network of camera traps. 
In another case, building real-time estimates of salmonid escapement requires teams of field ecologists working in shifts to watch streams of sonar data 24 hours a day. 
The challenge is even greater for taxa that are studied by trapping specimens, such as beetles and other insects.
Entomologists can collect thousands of beetles in a few days, but it may require months or years for a suitable expert to exhaustively identify all of the specimens to the species level. 
 
Machine learning methods can significantly accelerate the processing and analysis of large repositories of raw data \cite{iNat, Wildlife_Insights, LILA, berger2017wildbook, GBIF}, which can increase the speed and geographic scope of ecological analysis. 
For instance, ongoing collaborations between machine learning researchers and ecologists have lead to tremendous progress in automating species identification from images in community science data \cite{van2018inaturalist, aodha2019presenceonly} and camera trap data \cite{beery2019efficient, Wildlife_Insights}.
However, unfamiliar ecological concepts and terminology can present a barrier to entry for many computer scientists who might otherwise be interested in contributing to ecological problems. 
This is particularly true for more involved ecological problems which may not fit neatly into existing machine learning paradigms. 

One such area is \textbf{species distribution modeling} (SDM): using species observations and environmental data to estimate the geographic range of a species.\footnote{We will use the term ``species distribution modeling" throughout this document, though sometimes the closely related term ``ecological niche modeling" would be more appropriate \cite{peterson2012species}.}
This problem has received significant attention from ecologists and statisticians, and there has been increasing interest in machine learning methods due to the large amounts of available data and the highly complex relationships between species and their environments. 
This document is meant to serve as an easy entry point for computer scientists interested in SDM. 
In particular, we aim to highlight the exciting technical challenges posed by SDM while also emphasizing the needs of end-users to encourage ecologically meaningful progress.
Our hope is that this document can serve as a quick resource for computer science researchers interested in getting started working on conservation and sustainability applications.

The rest of this work is organized as follows. 
In Section \ref{sec:distributions} we discuss different ways to represent the distribution of a species.
We discuss species distribution modeling in Section \ref{sec:sdm} and we consider other related ecological modeling problems in Section \ref{sec:other_models}. 
In Section \ref{sec:pitfalls} we point out pitfalls and challenges in SDM.
Finally, we provide pointers to available data (Section \ref{sec:datasets}) and discuss open problems (Section \ref{sec:open_problems}).

\section{Representing the distribution of species}\label{sec:distributions}

The distribution of a species is typically represented as a \emph{map} which indicates the spatial extent of the species. 
These maps can be created in a variety of ways, ranging from highly labor-intensive expert range maps to fully automatic species distribution models. 
We show four examples in Fig.~\ref{fig:distributions_comparison}.
In this section we give a high-level overview of three important sources of maps: raw species observation data, predictions from statistical models, and expert knowledge. 

\subsection{Raw species observation data.}

Any representation of the distribution of a species begins with some sort of \emph{species observation data}. 
In general, species observation data consists of records indicating whether a species is present or absent at certain locations.
Species observation data can take many forms -- see Table \ref{tab:data_collection_methods} for examples. 
Species observation data falls into two general categories: \textbf{presence-only} data reports known sightings, or occurrences, of a species, while \textbf{presence-absence} data also provides information on where a species did not occur. 
Data collection strategies define whether absence data will be available. 
For instance, iNaturalist collects opportunistic imagery of species from community scientists, which produces presence-only species observations. 
On the other hand, eBird uses species \emph{checklists} where \textit{all} bird species seen and/or heard within a time span at a given location are reported. 
Since exhaustive reporting is expected from observers, any bird species not reported is assumed to be absent. 
In this sense, checklists are treated as presence-absence data.

One of the simplest ways to convey the distribution of a species is to simply show all of the locations where the species is known to be present or absent on a map.
However, this sort of highly simplified ``species distribution" is not able to make any predictions about whether a species might be present or absent at locations which have not been sampled.

\subsection{Statistical models.}

To create species distributions that can extrapolate beyond sampled locations, we can pair species observations with collections of environmental characteristics (altitude, land cover, humidity, temperature, etc.) and fit statistical models that use the environmental characteristics to predict species presence or absence. 
These models can make predictions at any place and time for which these environmental characteristics are known. 
Species distribution models fall into this category, and are our focus throughout this document. 

\subsection{Expert range maps.}

\begin{figure}
    \centering
    \includegraphics[width=8cm]{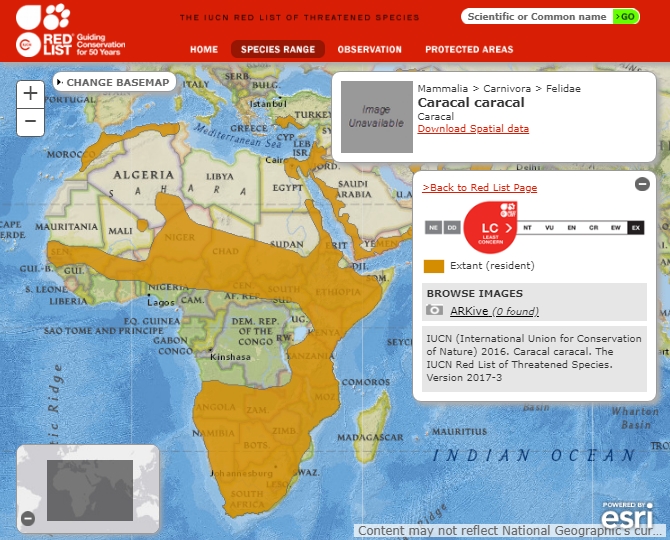}
    \caption{The International Union for Conservation of Nature (IUCN) publishes expert range maps for many species, particularly those on their ``Red List of Threatened Species" \cite{vie2009iucn}. Here we show the IUCN Range Map for the \emph{Caracal caracal} \cite{caracal}.}
    \label{fig:range_map}
\end{figure}

Species range maps have traditionally been heavily influenced by the individual scientists who study those species. These maps are often based on a complex combination of heterogeneous information sources, including personal observations, understanding of the species' habitat preferences, local knowledge/reports, etc.
From our discussions with practitioners, we find that these \emph{expert range maps} (ERMs) are often the most trusted source of distribution information.
Perhaps the most widely-known expert range maps are those published by IUCN \cite{red2018mapping} as part of their \textit{Red List} of vulnerable and endangered species. 
An example of the IUCN range map for the \emph{caracal} can be seen in Fig.~\ref{fig:range_map}. 
Studies have shown both agreement \cite{alhajeri2019high} and disagreement \cite{hurlbert2005disparity, graham2006comparison} between ERMs and species observation data.
Expert range maps have also been found to be highly scale-dependent, tending to overestimate the occupancy area of individual species and ranges < 200km \cite{hurlbert2007species}. 
It is important to note that ERMs come in many forms, from hand-drawn maps to data-driven maps that are slightly refined by experts. 
In the latter case, ERMs are partially based on species observation data, so the two cannot be treated as independent sources. 
As we will discuss in more detail in Section \ref{eval}, the lack of a solid ``ground truth'' information about the true underlying distribution of species across space and time makes it difficult to analyze the accuracy of any species distribution model, including those drawn by experts.

\section{Species Distribution Models}\label{sec:sdm}

The terminology in this area can be confusing, so we will start with a definition and a few clarifications. 

\textbf{Intuitive definition.}
A species distribution model is a function that uses the characteristics of a location to predict whether or not a species is present at that location. 
This can be understood as a supervised learning problem.
The input is a vector of environmental characteristics for a location and the output is species presence or absence. 
In principle one could use almost any classification or regression technique as the basis for an SDM. 

\textbf{Formal definition.}
The key components of a simple species distribution modeling pipeline are: (1) species observation data, (2) a method for encoding locations, and (3) a function which maps location encodings to predictions. 
Formally, we define these components as follows:
\begin{enumerate}
    \item A dataset of species observations. This is a collection of records indicating that a species is present or absent at given location and time. We write this as $\{ (\mathbf{x}_i, y_i) \}_{i=1}^N$ where $\mathbf{x}_i \in \mathcal{X}$ is a spatiotemporal location and $y_i \in \{0,1\}$ indicates presence ($1$) or absence ($0$). The spatiotemporal domain $\mathcal{X}$ is typically something like $\mathcal{X} = [0,180) \times [0,360) \times [0,1)$ which encodes global longitude and latitude as well as the time of year. 
    \item A location representation $h: \mathcal{X} \to \mathcal{Z} \subset \mathbb{R}^k$. This is typically a simple ``look-up" operation, where $\mathbf{x} \in \mathcal{X}$ is cross-referenced with $k$ pre-defined geospatial data layers to produce a vector of location features $h(\mathbf{x}) \in \mathbb{R}^k$. 
    That is, $h(\mathbf{x})$ is a representation of the location $\mathbf{x}\in \mathcal{X}$ in some environmental feature space.  
    \item A model $f_\theta: \mathcal{Z} \to [0,1]$ where $\theta$ is a parameter vector. The goal is to find parameters $\theta$ of $f$ so that $f_\theta(h(\mathbf{x})) = 1$ when the species is present and $f_\theta(h(\mathbf{x})) = 0$ otherwise. This is usually framed as a supervised learning problem on the dataset $\{ (h(\mathbf{x}_i), y_i) \}_{i=1}^N$.
\end{enumerate}
Note that this is a streamlined formalization meant to capture the essence of SDM. While there are many variants in practice, almost any species distribution modeling will include these core concepts.

\begin{figure*}
    \centering
    \includegraphics[width=0.95\textwidth]{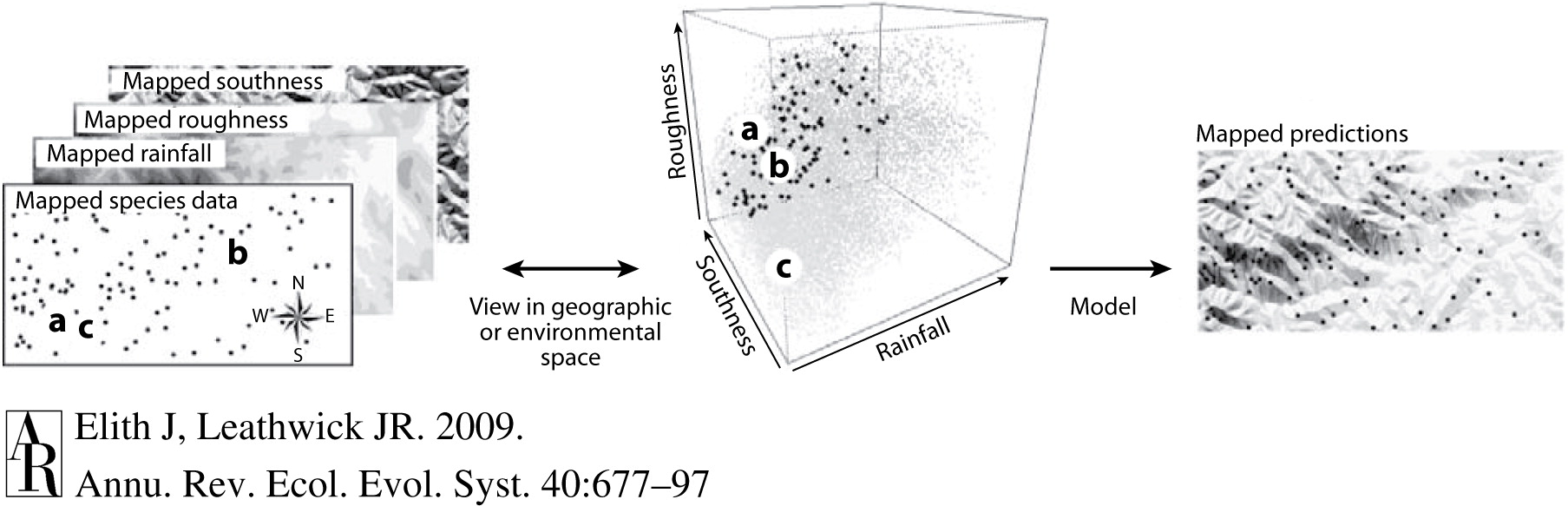}
    \caption{\textbf{Geographic vs. environmental space.} Observation data can be associated with a geographical location, or mapped into a feature space based on environmental covariates. Most SDMs operate under the assumption that with the right set of \textit{environmental variables} and an appropriate model, one could use environmental characteristics to map species distribution. Figure reproduced with permission, originally published in \cite{elith2009species}.}
    \label{fig:geo_vs_env}
\end{figure*}

\textbf{What does an SDM actually predict?}
An SDM takes as input a vector of environmental features and predicts a numerical score (usually between $0$ and $1$) for a location. An important distinction to note regarding SDMs is \textit{geographic space} vs. \textit{environmental space}, elucidated in Fig.~\ref{fig:geo_vs_env}. 
This score is often interpreted as a prediction of habitat suitability. 
Typically the score \emph{may not} be interpreted as the probability a species is present.
Note that here we are only considering presence vs. absence - predicting species \emph{abundance} is a more challenging problem, which we discuss in Section \ref{abundance}.

\textbf{How is an SDM used?}
The most common end product is a map of the SDM predictions, which is produced by simply visualizing the SDM predictions across an area of interest. 
Binary predictions can be obtained by applying a threshold to the continuous predictions of the SDM. 

\subsection{A brief history of species distribution modeling}

Early predecessors for SDM include qualitative works that link patterns within taxonomic groups to environmental or geographic factors, such as Joseph Grinnel's 1904 study of the distribution of the chestnut-backed chickadee \cite{grinnell1904origin}, among others \cite{murray1866geographical,schimper1903plant,whittaker1956vegetation, macarthur1958population}. 

Modern SDMs are primarily statistical models fit to observed data. Early quantitative approaches used multiple linear regression and linear discriminant function analyses to associate species and habitat \cite{capen1981use, stuffer2002linking}. The application of generalized linear models (GLMs) \cite{nelder1972generalized, austin1985continuum} provided more flexibility by allowing non-normal error distributions, additive terms, and nonlinear relationships. The explosive proliferation of large ``presence-only" datasets (see Table \ref{tab:data_collection_methods}) in recent years has led to the development of new modeling approaches to SDMs such as the popular ``Maximum Entropy Modeling" (MaxEnt) approach \cite{phillips2006maximum} with roots in point process modeling \cite{renner2013equivalence}.

The first modern SDM computing package, BIOCLIM, was introduced in 1984 on the CSIRO network \cite{busby1991bioclim, booth2014bioclim}. This package took observation information, such as the species observed, location, elevation, and time, and used them to determine what environmental variables correlated with that species' occurrence. These variables were then used to map possible distributions of the species under consideration. Climate interpolation techniques developed for BIOCLIM are the basis of the existing WorldClim database \cite{fick2017worldclim} and are still widely used in SDMs today. Many different implementations of various SDM methods are now publicly available. We would like to highlight Wallace \cite{kass2018wallace}, which is a well-documented R implementation of historic and modern techniques.

As earth observation technology has improved, the scope of what is possible to include as an environmental covariate in a model has vastly increased. Improvements in weather monitoring systems gave access to high-temporal-frequency temperature, wind, and precipitation measurements. Recently, ecologists have turned to remote sensing imagery to estimate high-spatial-coverage ecological variables such as soil composition or density of sequestered carbon, as well as mapping land cover type across regions \cite{he2015will}. Modern SDM methods pair these covariate estimates with increasingly accurate global elevation maps, and selected high-quality but sparse in-situ measurements \cite{ pradervand2014very, lassueur2006very}.

Several excellent, detailed reviews of SDMs have been published within the ecology community \cite{elith2009species, guisan2000predictive, stuffer2002linking, guisan2005predicting, richards2007distribution, schroeder2008challenges}. We direct the reader to the excellent summary by Elith and Leathwick \cite{elith2009species}.

\subsection{Covariates for species distribution modeling} \label{sec:covariates}

In this section we discuss several environmental characteristics (often called \emph{covariates}) that can be used for species distribution modeling. 
Here we are focused on describing the different categories of covariates -- details on specific covariate datasets are available in Section \ref{sec:datasets}.  Some of the covariates we discuss are widely used in the species distribution modeling literature, while others are more recent or speculative. It is also important to keep in mind that many covariates are themselves based on sophisticated predictive models due to the cost of densely sampling any property of the earth's surface. 

\subsubsection{Climatic variables.} 

Temperature and precipitation are critical characteristics of an ecosystem. Perhaps the most commonly used climate dataset for SDM is the WorldClim bioclimatic variables \cite{fick2017worldclim} dataset, which provides 19 climate-related variables averaged over the period from 1970 to 2000 at a spatial resolution of around 1km$^2$. We show a few examples of variables from this dataset in the top row of Fig.~\ref{fig:env_vars_example}.

\subsubsection{Pedologic (soil) variables.} 

\begin{figure}
    \centering
    \includegraphics[width=\textwidth]{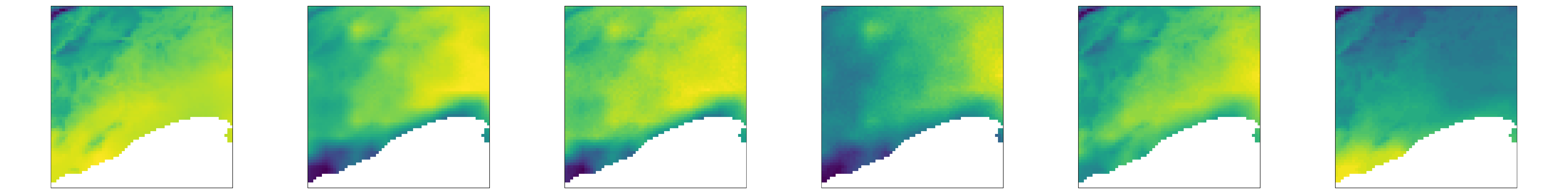} \\
    \includegraphics[width=\textwidth]{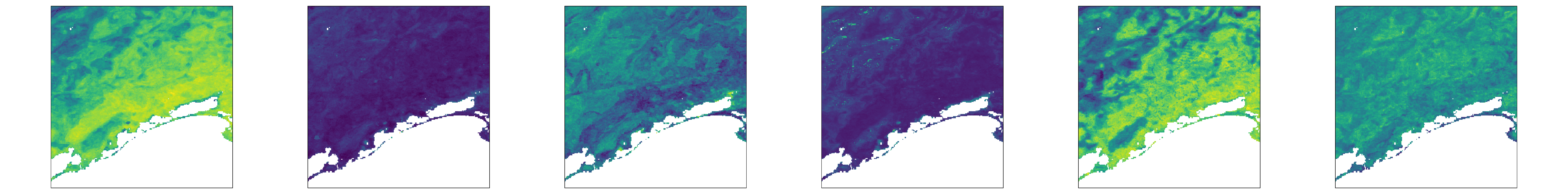}
    \caption{
    Visualizations of some of the bioclimatic variables (top row: \texttt{bio\_1} - \texttt{bio\_6} from left to right) and pedologic variables (bottom row: \texttt{orcdrc}, \texttt{phihox}, \texttt{cecsol}, \texttt{bdticm}, \texttt{clyppt}, \texttt{sltppt} from left to right) provided for the GeoLifeCLEF 2020 competition \cite{cole2020geolifeclef}.
    The area shown in each image is approximately 64~km$^2$ centered in Montpellier, France.
    While we visualize each environmental variable as a 2D raster, most species distribution modeling methods are only compatible with relatively low-dimensional vectors of environmental variables (not ``stacks" of 2D patches). 
    As is typical in a collection of covariates, we see that the pedologic variables have a different resolution than the bioclimatic variables.
    } 
    \label{fig:env_vars_example}
\end{figure}

Soil characteristics are intimately related to the plant life in an area, which naturally influences the entire ecosystem. 
One example of a comprehensive pedologic dataset is SoilGrids250m \cite{hengl2017soilgrids250m}, which consists of soil properties like pH, density, and organic carbon content at a 250m$^2$ resolution globally. We show a few examples of variables from this dataset in the bottom row of Fig.~\ref{fig:env_vars_example}.

\subsubsection{Vegetation indices.} 

A \emph{vegetation index} (VI) is a number used to measure something about the plant life in an area, and is typically computed from remote sensing data like satellite imagery. 
Many different VIs have been proposed.
A review paper published in 1995 discussed 40 different vegetation indices that had been developed by different researchers \cite{bannari1995review}. 
One of the most popular examples is the \emph{normalized difference vegetation index} (NDVI).
If a remote sensing image includes the red and near-infrared (NIR) bands, then the corresponding NDVI image can be computed by applying the formula
\begin{align}
    \mathrm{NDVI} &= \frac{\mathrm{NIR} - \mathrm{Red}}{ \mathrm{NIR} + \mathrm{Red}}
\end{align}
independently at each pixel.
NDVI is meant to indicate the presence of live green plants.  
From a computer vision perspective, these VIs are essentially hand-designed features for remote sensing data. 

\subsubsection{Land use / land cover.}

The term \emph{land cover} refers to the physical terrain at a location, while the closely related term \emph{land use} tends to emphasize the function of a location. 
For instance, an area with the land cover label ``dense urban" may have a land use label like ``school" or ``hospital." 
We provide an example in Fig.~\ref{fig:land_cover}, which shows RGB imagery and land cover from two different sources for the same 1km$^2$ area. 
It is not obvious what the best label set would be for species prediction, but practically speaking many of the available land use / land cover datasets are focused on relatively coarse categories related to agriculture, natural resources, or urban development. 
For instance, the U.S. National Land Cover Database assigns one of 20 land cover classes to every 30m$^2$ patch of land in the United States at a temporal resolution of 2-3 years \cite{homer2015completion}.
The classes cover various general habitat types (water, snow, developed land, forests...) but are not tuned for species prediction in particular.

\subsubsection{Measures of human influence.} 

Humans have had a profound impact on the natural world, so it is reasonable to include measures of human influence as environmental characteristics.
For instance, the Human Influence Index \cite{sanderson2002human} uses eight factors (human population density, railroads, roads, navigable rivers, coastlines, nighttime lights, urban footprint, and land cover) to compute a score that is meant to quantify how much an environment has been reshaped by humans.

\subsubsection{Remote sensing imagery.}

\begin{figure*}
    \centering
    \includegraphics[width=0.6\textwidth]{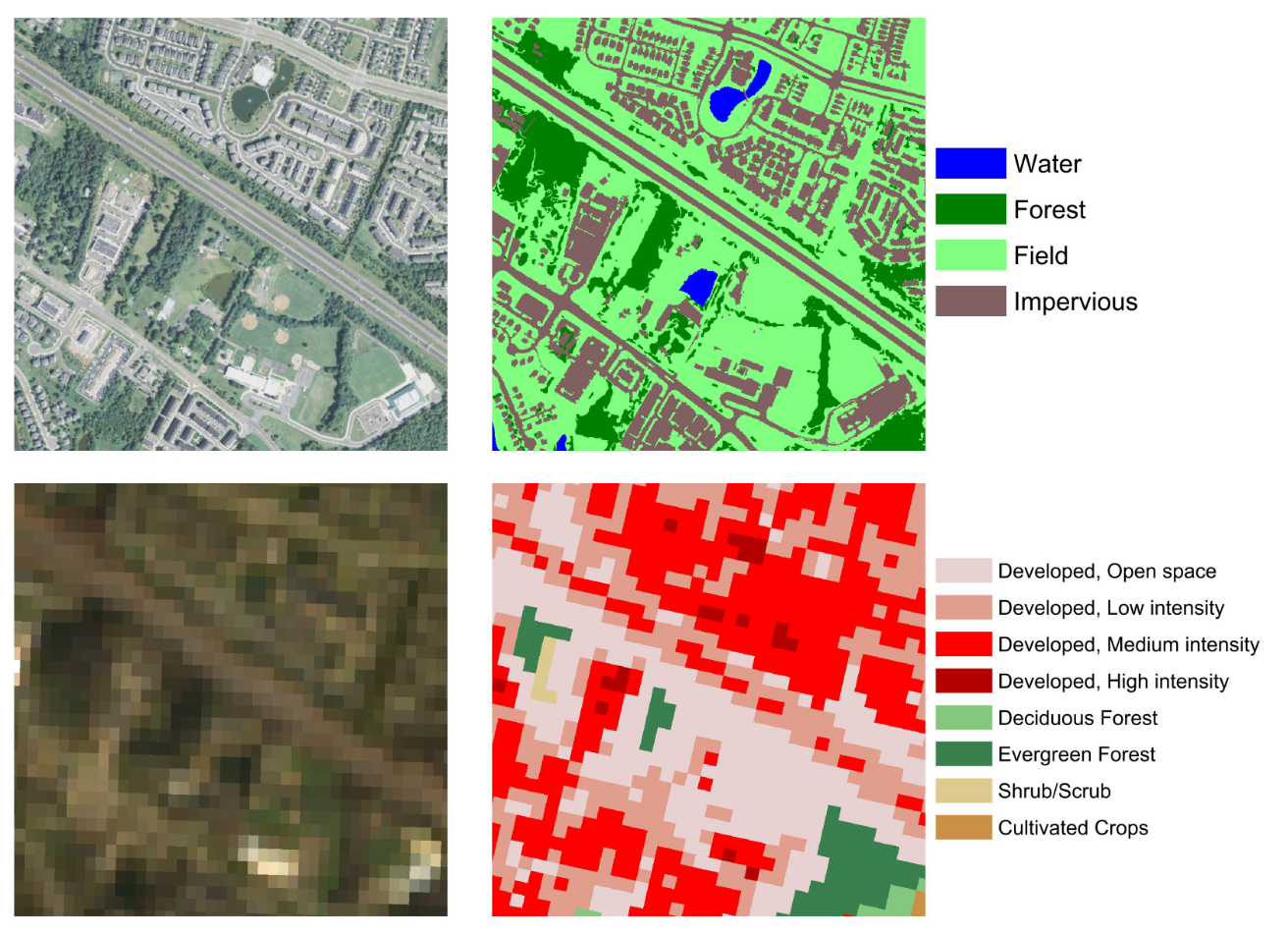}
    \caption{
    RGB imagery (left column) and land cover maps (right column) from two different remote sensing sources covering the same 1km$^2$ area, from \cite{robinson2019large}. RGB imagery is manually or semi-automatically annotated to produce the land cover labels. 
    As this example demonstrates, the set of land cover labels can vary depending on the organization doing the labeling. Figure reproduced with permission, originally published in \cite{robinson2019large}. 
    }
    \label{fig:land_cover}
\end{figure*}

Imagery collected by satellites, planes, or drones can provide substantial information about an environment. 
To start with, we note that vegetation indices, land cover, land use, and many measures of human influence are all derived from some form of overhead imagery like that in Fig.~\ref{fig:land_cover}.
In addition, there may be more abstract patterns that can be extracted using modern computer vision techniques like convolutional neural networks.
Research on the use of raw overhead imagery (instead of derived products) for SDM is in its early stages \cite{tang2018multi, cole2020geolifeclef, dobrowski2008mapping}. 

\subsection{Properties of species distribution models}\label{sec:sdm_properties}

In this section we describe important properties that can be used to categorize species distribution models. Any particular species distribution model may or may not have any of these properties. The categories we describe are in general nested or overlapping, not mutually exclusive.

\subsubsection{Presence only vs. presence-absence models.}

Species observation datasets may be either presence-absence or presence-only. 
While presence-only data is easier to collect, the are limitations on what can be estimated from such data \cite{hastie2013inference}.
Typically a species distribution model is designed to handle either presence-absence or presence-only data, though there is growing interest in developing methods that can use both \cite{gormley2011using, pacifici2017integrating, fletcher2019practical}.

\subsubsection{Single vs. multi-species models.}

Many SDMs are designed to model the distribution of a single species. 
This is in contrast to \emph{multi-species} models which are meant to capture information about several species. Many of the earlier models are single-species models \cite{phillips2006maximum, elith2009species}, though interest in multi-species models has grown over time \cite{hui2013mix, harris2015generating, norberg2019comprehensive}.

\subsubsection{Multi-species models: stacked vs. joint}

Multi-species SDMs can be classified as either \emph{stacked} or \emph{joint}.
In a \emph{stacked} model, a single-species SDM is fit for each species and the resulting maps are ``stacked" on top of one another to provide a multi-species map.
This approach is simple, but it cannot take advantage of patterns in how species co-occur. 
This is the motivation for \emph{joint} SDMs, in which the estimated distribution of each species also depends on occurrence data for other species.
Recent work has begun to systematically compare the results from stacked and joint species distribution models for different species and regions \cite{henderson2014species, norberg2019comprehensive, zurell2020testing}.

\subsubsection{Spatially explicit models.}

Typically species distribution models use environmental characteristics to make predictions about the presence or absence of species.
Such models represent a location in terms of these environmental features, so two different locations with the same environmental characteristics will lead to the same predictions, even though the two locations may be far apart.
Models that mitigate this concern by incorporating geographical location information directly are referred to as \emph{spatially explicit} \cite{domisch2019spatially} models.

\subsubsection{Occupancy models.}\label{Occupancy}

It is easier to confirm that a species is present than it is to confirm that a species is absent. 
One confident observation of a species suffices to confirm its presence at a given location.
However, failing to observe a species at a location does not suffice to prove absence, since the species could have been present but not observed.
\emph{Occupancy models} are meant to account for imperfect detection by modeling the probability that a species is present but unobserved at a given location conditional on the sampling effort that has been invested \cite{mackenzie2002estimating, bailey2014advances}.

\subsubsection{Understanding uncertainty and error.}

Species distribution models attempt to capture the behavior of a complex system from data, which is a challenging and error-prone process.
\cite{rocchini2011accounting} describes 11 sources of uncertainty and error in species distribution models, and groups them into two clusters: (i) uncertainty in the observation data itself and (ii) uncertainty due to arbitrary modeling choices.
\cite{dormann2008components} studies the effect of making different reasonable modeling choices on final projections of species distribution under different future climate scenarios.
Similarly, \cite{synes2011choice} considers the uncertainty introduced by the arbitrary choice of covariates while \cite{stoklosa2015climate} analyzes the effect of uncertainty in the values of the covariates themselves.
\cite{naimi2014positional} focuses on the effect of uncertainty in the location of species observations.
\cite{beale2012incorporating} reviews sources of uncertainty for different types of species distribution models, as well as best practices for minimizing uncertainty and methods for incorporating uncertainty directly into the model.

\subsection{Algorithms for species distribution modeling}

In this section we provide a high-level overview of the space of algorithms commonly used for species distribution modeling in the ecological community.
This section draws heavily from the organization of \cite{norberg2019comprehensive}, which is an excellent comparative study of different species distribution modeling techniques. We discuss several commonly used models, and note that the different methods can have very different properties, assumptions, and use cases.
Unlike some classes of algorithms, different species distribution modeling methods are generally not readily interchangeable.

\subsubsection{Presence-only methods.}

Perhaps the most popular approach for presence-only SDM is \emph{MaxEnt} \cite{phillips2006maximum}. 
We follow the description given in \cite{elith2011statistical}.
The basic idea is to estimate the probability of observing a given species as a function of the environmental covariates. 
The estimate is chosen to be (i) consistent with the available species observation data and (ii) as close as possible (in KL divergence) to the marginal distribution of the covariates. 
Criterion (ii) is necessary because there are typically many distributions that satisfy criterion (i).
Another simple approach for presence-only SDM is to introduce artificial negative observations called \emph{pseudonegatives} or \emph{pseudoabsences} based on some combination of domain knowledge and data. Once pseudonegatives have been generated, they are combined with the presence-only data and traditional presence/absence methods are applied. 

\subsubsection{Traditional statistical methods.}

Perhaps the most common methods in species distribution modeling are workhorse methods drawn from the statistics literature such as generalized linear models \cite{friedman2010regularization, foster2010analysis, wang2012mvabund, venables2013modern, ovaskainen2017make}.
Important special cases include logistic regression \cite{pearce2000evaluation} and generalized additive models \cite{wood2011fast}.
Some species distribution modeling algorithms are better thought of as general frameworks whose particular realization depends on the available data sources and modeling goals.
As an example, the Hierarchical Modeling of Species Communities (HMSC) framework \cite{ovaskainen2017make} minimally requires species occurrence data with corresponding environmental features.
The species occurrences are related to environmental features by a generalized linear model.
However, the framework can be extended to incorporate e.g. information on species traits and evolutionary history.

\subsubsection{Machine learning methods.}

The relationship between species and their environment is complex and may not satisfy traditional statistical assumptions such as linear dependence on covariates or i.i.d. sampling.
For this reason, machine learning approaches have also enjoyed considerable popularity in the species distribution modeling literature.
Examples include boosted regression trees \cite{elith2008working}, random forests \cite{cutler2007random}, and support vector machines \cite{drake2006modelling}. 
In addition, neural networks have been used for species distribution modeling since well before the deep learning era \cite{broomhead1988radial, ozesmi1999artificial, yen2004large, tirelli2009use}. 
Interest in joint species distribution modeling with neural networks has only grown as deep learning has come to maturity \cite{harris2015generating}.
Convolutional neural networks in particular have created a new opportunity: the ability to extract features from spatial arrays of environmental features \cite{chen2016deep,deneu2019evaluation} instead of using hand-selected environmental feature vectors.

\subsection{The challenge of evaluation}\label{eval}

How can we tell whether a species distribution model is performing well or not? 
The typical approach in machine learning is to use the model to make predictions on a held-out set of data and compute an appropriate performance metric by comparing the model predictions to ground-truth labels. 
But what is ``ground truth" for a species distribution model? 

\subsubsection{Notions of Ground Truth}

 We describe several common approaches to the challenging problem of how to evaluate SDMs in practice. For further detail, \cite{mouton2010ecological} provides an excellent discussion of different metrics for evaluating SDMs and the extent to which they are ecologically meaningful.

\textbf{Compare against presence-absence data.}  
Ideally, for each location, an expert observer would determine whether each species of interest is present or absent at that location.
Conducting this kind of survey for a single species in a limited area is expensive, and the survey would need to be repeated periodically to monitor change over time.
These exhaustive surveys quickly become extraordinarily expensive as we expand the number of species of interest or the geographic extent of the survey.
Even if the resources were available, the observations would have some degree of noise - in particular, confirming that a species is absent from an area can typically only be done up to some degree of certainty. 
(See the discussion of occupancy modeling in Section \ref{Occupancy}.)
For most species and most locations on earth, this sort of ideal ground truth data is just not available. 
However, this kind of evaluation is possible for select species and locations at sparse time points. 
For instance, \cite{elith2020presence} includes presence-absence data for 226 species from 6 parts of the world collected at various time points. 

\textbf{Compare against presence-only data.} Unfortunately, presence-absence data is often unavailable. We describe a few simple methods for comparing predictions against presence-only data along with their shortcomings.
\begin{itemize}
    \item False negative rate: how often are locations which are known to be positive predicted to be negative? The false negative rate measures whether the model is consistent with the observed positives, but does not assess the model's behavior at other points. 
    \item Top-$k$ classification accuracy: how often is the observed species among the $k$ most likely species under the model? However, there is not an obvious way to choose $k$. Moreover, for any fixed $k$ it is likely that some locations will have more than $k$ species while others will have fewer. 
    \item Adaptive top-$k$ classification accuracy: this is a variant of the top-$k$ classification accuracy that assumes that the number of species is $k$ on average, while allowing some locations to have more than $k$ species while others may have fewer. See \cite{cole2020geolifeclef} for details. Like standard top-$k$ classification accuracy, choosing $k$ may be difficult.
\end{itemize}
Note that adaptive top-$k$ and top-$k$ are both metrics for multi-species models, while the false negative rate can be computed for single species models as well.

\textbf{Compare against community science data.}
Community science projects like iNaturalist and eBird are generating species observation data at an extraordinary rate and frequency. iNaturalist alone generates millions of species observations per month \cite{inat50M}. 
However, the data produced by such projects can vary in terms of how easy it is to use and interpret depending on the sampling protocol \cite{kosmala2016assessing}. 
For instance, iNaturalist accepts presence-only observations, which allows the user base to scale broadly but limits the utility of the data for ground truthing. 
iNaturalist data tells us where different species have been observed by humans, but not where those species are either absent or present without human observation. 
eBird uses a more rigorous sampling protocol that records both presences and absences, but their observations are limited to birds.
The quality of these reports depends on the skill of the user at identifying all bird species they see or hear.
Citizen science data has been found to produce results similar to those from (coarse) professional surveys under the right circumstances \cite{higa2014mapping,tye2016evaluating,kosmala2016assessing}.

\textbf{Compare against expert range maps.}
Another possibility is to compare the model predictions against one or more range maps that are hand-drawn by experts (see Section \ref{sec:distributions}). 
However, this raises the question: how do we validate \emph{those} range maps?
A hand-drawn map may be biased by an individual's experience or by the data sources the expert prefers. 
In addition, it can be difficult to find a suitable expert to generate a map for every species of interest. 
Another challenging question relates to temporal progression: is each expert updating their maps according to the latest data?
If so, when was that data collected? 
The IUCN has a published set of standards for creating species range maps \cite{red2018mapping}, but not all creators of maps match these standards.

In addition, there is the methodological question of how one should evaluate a model against an expert range map, which is explored in \cite{mainali_hefley_ries_fagan_2020}. Approaches range from very qualitative (ask an expert whether the map looks reasonable to them) to very quantitative (compute a well-defined error metric between the SDM predictions and the expert range map). 
Important to note here, expert range maps are most often categorical, with hard boundaries drawn representing temporal categories like ``breeding'', ``non-breeding'', ``year-round'', etc. 
On the other hand, SDM predictions are often real-valued on $[0, 1]$ over both space and time. 
While continuous predictions can be converted to binary maps by applying a threshold, it can be unclear how to choose this threshold if a robust validation method is not available.

\textbf{Evaluation on downstream tasks.}
Instead of evaluating whether a species distribution model produces a faithful map of species presence, we may instead check whether it is useful for some other downstream task. For example, \cite{aodha2019presenceonly} builds a simple SDM and demonstrates that it improves accuracy on an image-based species classification task. However, it is certainly possible for an SDM to be useful whether or not it accurately reflects the true species distribution.

\subsubsection{Evaluation pitfalls}

Even when suitable ground truth data is available, there are some pitfalls that can hinder meaningful evaluation. 
In this section we discuss some of these pitfalls and make specific recommendations to the machine learning community for handling them. 

\textbf{Performance overestimation due to spatial autocorrelation.} 
In the machine learning community it is common to sample a test set uniformly at random from the available data. 
However, this strategy can lead to overestimation of algorithm performance for spatial prediction tasks since it is possible to obtain high performance on a uniformly sampled test set by simple interpolation \cite{roberts2016cross}. 
This effect is called \emph{spatial autocorrelation}. 
Similar concerns are relevant for evaluating camera trap image classifiers \cite{beery2018recognition}. 
For ecological tasks, it is important to evaluate models as they are intended to be used. In many cases, the more ecologically meaningful question is whether the model generalizes to novel locations, unseen in the training set. In these cases it is important to create a test set by holding out spatial areas. In other cases, the ecologist seeks to build a model that will perform accurately in the future at their set of monitoring sites. In these cases, instead of holding out data in space, we can split the data to hold out a test set based on time. A randomly sampled test set is not a good proxy for the use case of either scenario.

\textbf{Hyperparameter selection.} 
The performance of an algorithm typically depends on several hyperparameters. 
In the machine learning community these are set using cross-validation on held-out data. 
However, selecting and obtaining a useful validation set can be particularly challenging in SDM due to the data collection challenges described elsewhere. 
Recent work has also studied the sensitivity of SDMs to hyperparameters \cite{hallgren2019species} and developed techniques for hyperparameter selection in the presence of spatial autocorrelation \cite{schratz2019hyperparameter}.

\textbf{Spatial quantization.}
A natural first step when working with spatially distributed species observations is to define a spatial quantization scheme.
By ``binning" observations in this way, we can associate many species observations with a single vector of covariates. 
Additionally, spatially quantized data can be more natural from the perspective of many machine learning algorithms since the domain becomes discrete. 
However, the choice of quantization scheme (grid cell size) is difficult to motivate in a rigorous way. 
This is a problem because different quantization choices can result in vastly different outcomes - this is known as the \emph{modifiable areal unit problem} \cite{openshaw1984maup}. 
It is possible to cross-validate the quantization parameters, but only in those limited cases where there is enough high-quality data for this to be a reliable procedure. 
Furthermore, that process may be computationally expensive. 

\textbf{The long tail.}
Many real-world datasets exhibit a \emph{long tail}: a few classes represent a large proportion of the observations, while many classes have very few observations \cite{van2017devil, beery2018recognition}. 
Species observation data is no exception - for example, in the Snapshot Serengeti camera trap dataset \cite{swanson2015snapshot} there are fewer than 10 images of gorillas out out of millions of images collected over 11 years. 
This presents at least two problems.
The first problem is that standard training procedures will typically result in a model that perform well on the common classes and poorly on the rare classes. 
The second is that many evaluation metrics are averaged over all examples in the dataset, which means that the metric can be very high despite poor performance on almost all species.
It is much more informative to study the performance on each class or on groups of classes (e.g. common classes vs. rare classes). One common solution is to compute metrics separately for each class and then average over all classes to help avoid bias towards common classes in evaluation.

\subsubsection{Model trust}

Once a model has been built, the previously discussed challenges of model evaluation make it difficult to determine where, how much, and for how long a model is sufficiently accurate to be used. The accuracy needed may also vary by use case and subject species. In our discussions with ecologists, we find that this leads to a lack of trust in SDMs. What verification and quality control is needed to ensure a model is still valid over time? This is an open question, and an important one to answer if our models are to be used in the real world.

\section{Other types of ecological models}\label{sec:other_models}

Species distribution modeling is only one of many ways that ecologists seek to describe and understand the natural world. To give readers a sense of how SDM fits into the broader scope of ecological modeling, we provide a high-level overview of other common modeling tasks.

\subsection{Mechanistic models}

Mechanistic models make assumptions about how species depend on the environment or on other species. One example is to use an understanding of a plant's biology to predict the viable temperature range where the plant can grow \cite{sweeney1975vegetative}. Such models are useful but difficult to scale, as they require species-specific expert knowledge.
Our focus in this work is on \emph{correlative} species distribution models, which do not require mechanistic knowledge. 

\subsection{Abundance modeling}\label{abundance}

\emph{Abundance modeling} goes beyond species presence or absence, aiming to characterize the absolute or relative number of individuals at a given location. We define abundance and related concepts in Section \ref{sec:biodiv_measurement}.

\subsubsection{Population estimation} 

Population estimation is concerned with counting the total number of individuals of a species, typically within some defined area \cite{schnabel1938estimation}.
Population size is most frequently estimated using \emph{capture-recapture models}, which require the ability to distinguish between individuals of the same species. 
Traditionally this individual re-identification was based on physical tags or collars \cite{grimm2014reliability}, but some recent efforts have relied on the less invasive method of identifying visually distinctive features, such as stripe patterns or the contour of an ear \cite{berger2017wildbook}.

\subsubsection{Density estimation} 

Density estimation seeks to model \emph{spatial abundance}, the abundance of a species per unit area, to understand where a species is densely versus sparsely populated \cite{rowcliffe2008estimating, verberk2011explaining}. 

\subsubsection{Data collection procedures for abundance}

As mentioned above, capture-recapture requires an individual to be re-identifiable. In the absence of the ability to re-identify individuals, several other data collection procedures are used. One that is frequently used for insects and fish populations is the \emph{harvest method}, where individuals are collected in traps which are open for a set amount of time and then counted \cite{pope2010methods, seibold2019arthropod}. Sampling strategies for other taxa include:

\begin{itemize}
    \item \textbf{Quadrat sampling.} A \emph{quadrat} is a fixed-size area where species are to be sampled. 
    Within the quadrat, the observer exhaustively determines the occurrence and relative abundance of the species of interest.
    Quadrat sampling is most commonly used for stationary species like plants. 
    The observer will sample quadrats throughout the region of interest to derive sample variance and conduct further statistical analysis \cite{hanley1978comparison}.
    \item \textbf{Line intercept sampling.} A \emph{line intercept} or \emph{line transect} is a straight line that is marked along the ground or the tree canopy, and is primarily used for stationary species \cite{heady1959comparison}. The observer proceeds along the line and records all of the specimens intercepted by the line. Each transect is regarded as one sample unit, similar to a single quadrat.
    \item \textbf{Cue counting.} Cue counting is based on observing cues or signals that a species is nearby, such as whale or bird calls. It is used primarily for species that are underwater or similarly difficult to sight \cite{marques2011estimating}.
    \item \textbf{Distance sampling.} \emph{Distance sampling} refers to a class of methods which estimate the density of a population using measured distances to individuals in the population \cite{buckland2005distance}. 
    Distance sampling can be added to line transects in order to incorporate specimens that are off the transect line but still visible.
    Appropriately calibrated camera traps can also benefit from distance sampling \cite{rowcliffe2008estimating}.
    \item \textbf{Environmental DNA (eDNA) sampling.} Samples of water or excrement collected in the field can be sequenced to provide species identifications. The ratios of environmental DNA for each species can be used to estimate abundance \cite{lodge2012conservation, valentini2016next}.
\end{itemize}

Each of these procedures produces different types of data, and each method comes with its own innate collection biases. These biases can add to the challenge of evaluating ecological models, as discussed in Section \ref{eval}.

\subsection{Biodiversity measurement and prediction}\label{sec:biodiv_measurement}

While it is important to understand the distribution of particular species, in many cases the ultimate goal is to understand the health of an ecosystem at a higher level.
\emph{Biodiversity} is a common surrogate for ecosystem health, and there are many different ways to measure it \cite{whittaker1960vegetation, jost2006entropy, jost2007partitioning}.
In this section we define and discuss several biodiversity metrics and related concepts.
Note that some sources give different definitions than those presented here, so caution is warranted.

We now define some preliminary notation.
We let $R$ denote an arbitrary spatial unit such as a country.
Many biodiversity metrics are computed based on a \emph{partition} of $R$ into $N$ sub-units, which we denote by $\{R_i\}_{i=1}^N$.
The choice of partition can have a significant impact on the value of some metrics, but for the purposes of this section we simply assume a partition has been provided.

\textbf{Species richness.}
The species richness of $R$ is the number of unique species in $R$, which we write as $S(R)$. 

\textbf{Absolute abundance.} 
The absolute abundance of species $k$ in $R$ is the number of individuals in $R$ who belong to species $k$. We write this as $A_k(R)$.

\textbf{Relative abundance.} 
The relative abundance of species $k$ in $R$ is the fraction of individuals in $R$ who belong to species $k$, which is
\begin{align}
    p_k(R) = \frac{A_k(R)}{\sum_{j=1}^{S(R)} A_j(R)}.
\end{align}
Since $\sum_{j = 1}^{S(R)} p_j(R) = 1$ and $p_j(R) \geq 0$ for all $j\in \{1,\ldots,S(R)\}$, the vector of relative abundances $\mathbf{p}(R) = (p_1(R),\ldots,p_{S(R)}(R))$ forms a discrete probability distribution.
The species richness can then be alternately defined as the support of this distribution, given by 
\begin{align}
    S(R) = |\{j \in \{1,\ldots, S(R)\}: p_j(R) > 0\}|.
\end{align}
Of course we can replace $p_j$ with $A_j$ everywhere and get an identical quantity.

\textbf{Shannon index.}
The Shannon index of $R$ is the entropy of the probability distribution $\mathbf{p}(R)$, so
\begin{align}
    H(\mathbf{p}(R)) = - \sum_{j=1}^{S(R)} p_j(R) \log p_j(R).
\end{align}
The Shannon index quantifies the uncertainty involved in guessing the species of an individual chosen at random from $R$.
Sometimes $H$ is instead written as $H'$, and sometimes the argument is written as $R$ instead of $\mathbf{p}(R)$. 

\textbf{Simpson index.}
The Simpson index of $R$ is the probability that two individuals drawn at random from the dataset (with replacement) are the same species, and is given by 
\begin{equation}
    \lambda(R) = \sum_{i=1}^{S(R)} p_i^2.
\end{equation}

\textbf{Alpha diversity.}
The alpha diversity of $R$ is the average species richness across the sub-units $\{R_i\}_{i=1}^N$, given by
\begin{align}
    \alpha(R) = \frac{1}{N}\sum_{i=1}^N S(R_i).
\end{align}

\textbf{Gamma diversity.}
The gamma diversity of $R$ is defined as 
\begin{align}
    \gamma(R, q) &= \left( \sum_{j=1}^{S(R)} p_j^q \right)^{1/(1-q)}
\end{align}
where $q \in [0, 1) \cup (1, \infty)$ is a weighting parameter \cite{jost2006entropy}.
Note that gamma diversity is also commonly denoted by $^\gamma D_q(R)$.
There are several interesting special cases:
\begin{itemize}
    \item If $q=0$ then gamma diversity reduces to species richness i.e. $\gamma(R, 0) = S(R)$. 
    \item Gamma diversity is also related to the Shannon index, since $\lim_{q\to 1} \gamma(R,q) = \exp H(\mathbf{p}(R))$\cite{jost2006entropy}.
    \item If $q = 2$ then gamma diversity reduces to the inverse of the Simpson index i.e. $\gamma(R, 2) = 1/\lambda(R)$. 
\end{itemize}

\textbf{Beta diversity.}
The beta diversity of $R$ is meant to measure the extent to which sub-units $R_i$ are ecologically differentiated. This can be interpreted as a measure of the variability of biodiversity across sub-regions or habitats within a larger area. It is defined as
\begin{equation}
\beta(R, q) = \frac{\gamma(R, q)}{\alpha(R)}
\end{equation}
where $q$ is the same weighting parameter we say in the definition of gamma diversity \cite{tuomisto2010diversity, jost2006entropy}. 
Beta diversity quantifies how many sub-units there would be if the total species diversity of the region $\gamma$ and the mean species diversity per sub-unit $\alpha$ remained the same, but the sub-units had no species in common.

\section{Common challenges and risks}\label{sec:pitfalls}

\subsection{Differences in tools}

R is the dominant coding language in ecology and statistics, but Python is dominant in machine learning. This language barrier limits code sharing, which in turn limits algorithm sharing. It is also important to note that some machine learning models are extremely computationally demanding to train, and some ecologists may not have access to the necessary computational resources.

\subsection{Differences in ideas and terminology}

Differences in concepts and terminology can make it difficult for machine learning practitioners to find and read relevant work from the ecology community (and vice-versa). 
However, there is a growing body of interdisciplinary work which brings ecologists and computer scientists together \cite{AI_Animal_ReID,AI_CVWC_ICCV,AI_ESA}. It is important for computer scientists working in this area to establish ties with ecologists who can help them understand how to make ecologically meaningful progress.

\subsection{Combining data sources}

Species observation data is collected according to many different protocols, which means that effectively combining different data sources can be nontrivial \cite{pacifici2016integrating,koshkina2017integrated,miller2019recent,gelfand2019preferential}. For instance, observations collected in a well-designed scientific survey have significantly different collection biases from observations collected via iNaturalist. Handling these biases in a robust, systematic way can be quite challenging, particularly for large collections of data encompassing thousands of different projects, each with their own sampling strategies. In many cases, understanding the protocols used for a specific data collection project within a larger repository requires one to delve into the literature for that project.
However, for many projects there do not exist accessible, standardized definitions or quantitative analysis of bias.

\subsection{Black boxes, uncertainty, and interpretability}

Machine learning models are frequently ``black boxes," meaning that it is difficult to understand how a prediction is being made. Ecologists are accustomed to models that are simpler to inspect and analyze, where they can confidently determine what factors are most important and what the effect of different factors might be. Because the results of ecological models are used to drive policy, being able to interpret how a model is making predictions and avoid inaccuracies due to overfitting is important. This is closely related to trust (or lack thereof) in model outputs and the need for uncertainty quantification, particularly in scenarios where models are being asked to generalize to new locations or forward in time.

\subsection{Norms surrounding data sharing and open sourcing in ecology}

Computer science has benefited from strong community norms promoting public data and open-sourced code. 
One consequence of this shift is that it is easy for computer scientists to take data for granted and to be frustrated when a scientist is unwilling to share their data publicly.
However, it is important to remember that in some fields data can be extremely expensive to collect and curate. The cost of the hardware, travel to the study site, and the time needed to place the sensors and maintain the sensor network quickly adds up. Add to this the number of hours it takes for an expert to process and label the data so that it is ready for analysis, and it is easy to see why a researcher would want to publish several papers on their hard-won data before sharing it publicly. 
On the other hand, public datasets like those hosted on LILA.science \cite{LILA} have clear benefits for the community such as promoting reproducible research. 
Properly attributing data to the researchers who collected it (e.g. through the use of ``DOIs for datasets" \cite{robertson2019training}) could encourage more open data sharing in ecology.
Data sharing norms are changing and many researchers are now happy to share their data and are pushing for more open data practices \cite{powers2018open,reichman2011challenges}, but it is important to be aware of this cultural difference between computer science and other fields. 

\subsection{Model handoffs, deployment, and accessibility}

Once a machine learning method has been rigorously evaluated and found to be helpful, it is important to ensure these techniques are accessible to those who can put them to good use. In computer science, we have a culture of "open code, open data" which means that for most papers, all of the data and code is publicly available. However, ecologists may be less familiar with machine learning packages like PyTorch and TensorFlow, and may not have access to the computational resources required to train models on their data. If a method is to have real impact for the ecology community, it is important to provide models and code in a format that is accessible to end-users and well-documented. If the model is meant to become an integral part of an ecology workflow, plans for model maintenance and upkeep should be discussed.

\subsection{Sensitive species}

It is common for ecologists to obfuscate geolocation information before publishing any data containing rare or protected species to avoid poaching or stress from ecotourism. However, it is unclear whether obfuscation of GPS signal is sufficient to obscure the location of a photograph. 
It may be that a better solution is to remove any photos containing sensitive species, or to restrict sensitive access to a list of verified members of the research community. Second, the  obfuscation distance of GPS location in published datasets might have a large effect on the accuracy of an SDM or other ecological model, particularly when both the training and validation data have been obfuscated. This obfuscation will further effect the reproducibility of a study, as results with or without obfuscation might be quite different.

\section{What data is available and accessible?} \label{sec:datasets}

There is an increasing number of publicly available ecological datasets that can be used for model training and evaluation. In this section we provide a few useful data sources as a starting point. 
We make a distinction between ``analysis-ready" datasets which package species observations and covariates together and other data sources which can be combined to produce analysis-ready datasets. 

\subsection{Traditional analysis-ready datasets for multi-species distribution modeling}

\begin{itemize}
    \item The comprehensive SDM comparison in \cite{norberg2019comprehensive} uses five presence-absence datasets covering different species and parts of the world. 
    Each dataset has a different set of covariates (min 6, max 38) and a different set of species (min 50, max 242).
    The datasets are available for download on Zenodo  \cite{anna_norberg_2019_2637812}. 
    \item The recently released benchmark dataset \cite{elith2020presence} covers 226 species from 6 regions. Each region has a different set of covariates (min 11, max 13) and a different set of species (min 32, max 50).
\end{itemize}
 Note that many ``traditional" SDM datasets may not be large enough to train some of the more data-hungry machine learning methods. 

\subsection{Large-scale analysis-ready datasets for multi-species distribution modeling}

\begin{itemize}
    \item The GeoLifeCLEF datasets combine 2D patches of covariates with species observations from community science programs.
    The GeoLifeCLEF 2020 dataset \cite{cole2020geolifeclef} consists of 1.9M observations of 31k plant and animal species from France and the US, each of which is paired with high-resolution 2D covariates (satellite imagery, land cover, and altitude) in addition to traditional covariates.
    Previous editions of the GeoLifeCLEF dataset \cite{deneu2018location, botella2019overview} are also available, and are suitable for large-scale plant-focused species distribution modeling in France using traditional covariates.
    Note that all of the GeoLifeCLEF datasets are based on presence-only observations, so performance is typically evaluated using information retrieval metrics such as top-$k$ accuracy.
    \item The eBird Reference Dataset (ERD) \cite{munson2011ebird} is built around checklists collected by eBird community members. In particular, it is limited to checklists for which the observer (i) asserts that they reported everything they saw and (ii) quantified their sampling effort. This allows unobserved species to be interpreted as absences if sufficient sampling effort has been expended. The resulting presence/absence data is combined with land cover and climate variables. Unfortunately, the ERD does not appear to be maintained or publicly available as of November 2020. 
\end{itemize}

\subsection{Sources for species observation data}

\begin{itemize}
    \item The Global Biodiversity Information Facility (GBIF) \cite{GBIF} aggregates and organizes species observation data from over 1700 institutions around the world. We discuss a few specific contributors below. 
    \item iNaturalist \cite{iNat} is a community science project that has produced over 70 million point observations of species across the entire taxonomic tree. The data can be noisy as it is collected and labeled by non-experts.
    \item eBird \cite{eBird} is a community science project hosted by the Cornell Lab of Ornithology which has produced more than 77 million birding checklists. 
    These checklists provide both presence and absence, but absences can be noisy as it is possible the birder did not observe every species that was present at a given location.
    \item Movebank \cite{Movebank} is a database of animal tracking data hosted by the Max Planck Institute of Animal Behavior. It contains GPS tracking data for individual animals, covering 900 taxa and including 2.2 billion unique location readings.
\end{itemize}

\subsection{Sources for covariates}

Earth observation datasets and their derived products can be freely obtained from many sources, including the NASA Open Data Portal \cite{nasadata}, the USGS Land Processes Distributed Access Data Archive \cite{lpdaac}, ESA Earth Online \cite{esadata}, and Google Earth Engine \cite{earthengine}.
Also see the detailed discussion of covariates in Section \ref{sec:covariates}.

\subsection{Sources for training species identification models}

Species observation data can be produced by classifying the species found in geolocated images. Those who are interested in the species classification problem may be interested in the datasets below. 
\begin{itemize}
    \item The iNaturalist species classification datasets \cite{van2018inaturalist, van2021benchmarking} are curated species classification datasets built from research-grade observations in iNaturalist.
    \item LILA.science \cite{LILA, beery2018recognition, norouzzadeh2018automatically} hosts a number of biology-focused image classification datasets, including camera trap datasets covering diverse species and locations.
    \item The Fine-Grained Visual Categorization (FGVC) workshop \cite{FGVC} at CVPR hosts a number of competitions each year \cite{FGVC, beery2018iwildcam, beery2019iwildcam, beery2020iwildcam, beery2021iwildcam, van2018inaturalist, thapa2020plant, tan2019herbarium, mwebaze2019icassava} which focus on species classification and related biodiversity tasks.
\end{itemize}

\section{Open Problems}\label{sec:open_problems}

There are many open problems in SDM that may benefit from machine learning tools.
In this section we discuss a few of these problems which we find particularly interesting. 

\subsection{Scaling up, geospatially and taxonomically}

One of the main challenges in modern SDMs is scale.
This includes scaling up SDMs to efficiently handle large geographic regions~\cite{thuiller2008,kissling2018,jetz2019a}, many-species communities~\cite{norberg2019a,wilkinson2019b,pichler2020,tikhonov2020}, and large volumes of training data~\cite{merow2013,wilkinson2019a,tikhonov2020}. 
One particularly interesting question is whether jointly modeling many species could lead to SDMs which are significantly better than those based on modeling species independently.

\subsection{Incorporating ecological theory and expert knowledge}

There is a considerably amount of domain knowledge and ecological theory which would ideally be incorporated into SDMs~\cite{guisan2005a}.
This might include knowledge about species dispersal ~\cite{franklin2010,barve2011,miller2015,dimusciano2020}, spatial patterns of community composition ~\cite{damen2015,chen2017a,joseph2020a}, and constraints on species ranges (e.g. cliffs, water) ~\cite{fitzpatrick2009,ewers2010,miller2015,cooper2018}. 
Another area of significant interest is to factor in cross-species biological processes such as niche exclusion/competition ~\cite{wiens2011,poggiato2021}, predator/prey dynamics \cite{trainor2014,dormann2018,poggiato2021}, phylogenetic niche evolution \cite{pearman2008,gavin2014,chapman2017}, or models linked across functional traits \cite{pollock2012,clark2017a,vesk2021}. 
These types of ``domain-aware" algorithms are an active research area in the machine learning community \cite{bishop2013model, gu2016continuous, swischuk2019projection, doll2012ubiquity}.

\subsection{Fusing data}

A third open area of investigation centers on how to best incorporate and utilize data collected at different spatiotemporal scales or in heterogeneous formats. This includes combining presence-only, presence-absence, abundance, and individual data such as GPS telemetry data \cite{johnson2008,phillips2011,pacifici2017,fieberg2018}. It also includes multi-scale or cross-scale modeling~\cite{vaclavik2012,talluto2016}, such as microclimate niche vs. macroscale niche~\cite{lembrechts2019}, individual niche variance vs. species level niche variance\cite{fieberg2018}, and cross-scale ecological processes\cite{guillera-arroita2015,matthiopoulos2020}. Finally, it may also include models of temporal ecological processes, such as seasonal range shifts and migrations~\cite{thorson2016a,soriano-redondo2019}.

\subsection{Evaluation}

How should we compare competing models and decide which models to trust? 
Naturally, fair head-to-head evaluation of different models will be important \cite{araujo2006,elith2009,norberg2019comprehensive}. 
Future large-scale evaluations may require accounting for biases in species observation data \cite{vanderwal2009,ward2009,liu2013,fithian2015}, especially that which comes from community science projects. 
However, it is important to keep in mind that there is no single metric which makes one SDM better than another.
It may be important to understand how a model's predictions change under novel climate scenarios \cite{fitzpatrick2009,buisson2010,austin2011,liang2018} or different conservation policies \cite{sinclair2010,mcshea2014,eaton2018} or how well-calibrated the SDM predictions are \cite{araujo2006,grimmett2020}.
One promising avenue is to study models in increasing realistic simulation environments \cite{wisz2009,junior2018,meynard2019}, which allows for more comprehensive analysis. 
Many of these topics are directly related to active areas of machine learning research, such as generalization, domain adaptation, and overcoming dataset bias and imbalance \cite{koh2020wilds}.

\section{Conclusion}\label{sec:conclusion}

We have sought to introduce machine learning researchers to a challenging and important real-world problem domain. We have discussed common terminology and highlighted common pitfalls and challenges. To lower the initial overhead, we have inventoried some available datasets and common methods. We hope that this document is useful for any computer scientist interested in bringing machine learning expertise to species distribution modeling.

\begin{acks}
Our research for this paper included informational interviews with Meredith Palmer, Michael Tabak, Corrie Moreau, and Carrie Seltzer. Their insights into the unique challenges of species distribution modeling was invaluable. This work was supported in part by the Caltech Resnick Sustainability Institute and NSFGRFP Grant No. 1745301. The views expressed in this work are those of the authors and do not necessarily reflect the views of the NSF.
\end{acks}

\bibliographystyle{ACM-Reference-Format}
\bibliography{main}

\end{document}